\documentclass[11pt,a4paper]{article}
\usepackage[hyperref]{naaclhlt2019}
\usepackage{times}
\usepackage{latexsym}
\usepackage{amssymb} 
\usepackage{amsmath}
\usepackage{phonetic}
\usepackage{url}
\usepackage{xcolor}

\aclfinalcopy 
\def\aclpaperid{1338} 

\usepackage[disable]{todonotes}   
\definecolor{ForestGreen}{RGB}{0,0,0}

\newcommand{\note}[4][]{\todo[author=#2,color=#3,size=\scriptsize,fancyline,caption={},#1]{#4}} 
\newcommand{\katharina}[2][]{\note[#1]{Katharina}{green!40}{#2}}
\newcommand{\manuel}[2][]{\note[#1]{Manuel}{orange!40}{#2}}
\newcommand{\ozlem}[2][]{\note[#1]{Ozlem}{blue!40}{#2}}

\newcommand\StoS{Seq2Seq}

\title{Subword-Level Language Identification for Intra-Word Code-Switching}

\author{\parbox{\linewidth}{\centering
Manuel Mager{\rm\affmark[1]}, \"{O}zlem \c{C}etino\u{g}lu{\rm\affmark[1]} \and Katharina Kann{\rm\affmark[2]}} \vspace{.12cm}
\\
\affaddr{\affmark[1]Institute for Natural Language Processing,\\ 
University of Stuttgart, Germany}\\
\affaddr{\affmark[2]Center for Data Science, New York University, USA}\\
\affaddr{\texttt{\{manuel.mager, ozlem\}@ims.uni-stuttgart.de, kann@nyu.edu}}} 
\newcommand*{\affaddr}[1]{#1} 
\newcommand*{\affmark}[1][*]{\textsuperscript{#1}}

\date{}

\begin{document}
\maketitle
\begin{abstract}
Language identification for code-switching (CS), the phenomenon of alternating between two or more languages in conversations, has traditionally been approached under the assumption of a single language per token. However, if at least one language is morphologically rich, a large number of words can be composed of \textit{morphemes} from more than one language (intra-word CS). In this paper, we extend the language identification task to the subword level, such that it includes splitting mixed words while tagging each part with a language ID. 
We further propose a model for this task, which is based on a segmental recurrent neural network.
In experiments on a new Spanish--Wixarika dataset and on an adapted German--Turkish dataset, our proposed model performs slightly better than or roughly on par with our best baseline, respectively. Considering only mixed words, however, it strongly outperforms all baselines.
\end{abstract}

\ozlem{Manuel, there are a few comments within the text. Could you also make sure you cover the comments I gave on the printed version and the ones from our last meeting?}
\katharina{
\begin{itemize}
    \item LID means language identification by our definition. Manuel, you now also use it as language ID, which is confusing. Undo these abbreviations, please!
\end{itemize}}
\manuel{done}
\section{Introduction}
\label{sec:introduction}
In settings where multilingual speakers share more than one language, mixing two or more languages within a single piece of text, for example a tweet, is getting increasingly common \cite{grosjean:2010}. This constitutes a challenge for natural language processing (NLP) systems, since they are commonly designed to handle one language at a time. 

Code-switching (CS) can be found in multiple non-exclusive variants. For instance, sentences in different languages can be mixed within one text, or words from different languages can be combined into sentences. CS can also occur on the subword level, when speakers combine morphemes from different languages (\emph{intra-word CS}). This last phenomenon can mostly be found if at least one of the languages is morphologically rich. An example for intra-word CS between the Romance language Spanish and the  Yuto-Aztecan 
language Wixarika\footnote{Wixarika, also known as Huichol, is a polysynthetic Mexican indigenous language.} is shown in Figure \ref{tab:example1}. \katharina{Why is this saying tire instead of tired?}
\manuel{Because tired is the inflected form, and this only happens after adding the x+ morpheme.}\katharina{I think we could argue if this is a derived or inflected form. I would go with derived, since tired is a lemma for sure. Would you mind changing it?}

CS language identification (LID)
, i.e., predicting the language of each token in a text, has attracted a lot of attention in recent years (cf. \newcite{solorio2014overview,molina2016overview}). However, intra-word mixing
is mostly not handled explicitly: words with morphemes from more than one language are simply tagged with a \texttt{mixed} label. 

\begin{figure}[t!]
    \setlength{\tabcolsep}{3.6pt} 
    \begin{tabular}{p{.3cm} l l l l}
&\textbf{(a)}  & \textit{ne'iwa} & \textit{pecansadox\ibar} \\
&              & WIX & MIXED  \\
&              & my.brother & you-are.tired.PPFV  \\
& & &\\
    \end{tabular}

    \begin{tabular}{p{.3cm} l l l l l}
&\textbf{(b)} & \textit{ne'iwa} & \textit{pe}& \textit{cansado} &  \textit{x\ibar} \\
&             &WIX & WIX & ES &  WIX \\
&             &my.brother& you-are & tired &  PPFV \\
& & & & & \\
& &\multicolumn{3}{l}{{\color{ForestGreen}`My brother, you are tired.'}} 
    \end{tabular}
    \caption{Intra-word CS between Spanish and Wixarika, 
    \textbf{(a)} standard LID for CS, \textbf{(b)} our task. { \color{ForestGreen}PPFV stands for past perfective.}}
    \label{tab:example1}
\end{figure}

While this works reasonably well for previously studied language pairs, overlooking intra-word CS leads to a major loss of information for highly  {\color{ForestGreen} polsynthetic} \katharina{Check this; but I think we want to say polysynthetic, not agglutinative.} \manuel{I added polysynthetic because Turkish is agglutinative, so non of both typologies covers our languages}\katharina{This seems wrong! Let's discuss.}\manuel{ok.} languages. 
A mixed word is unknown for NLP systems, yet 
a single word contains much more information,
cf. Figure \ref{tab:example1} (b).
Furthermore, we find intra-word CS to be much more frequent for Spanish--Wixarika than for previously studied language pairs, such that it is crucial to handle it.




Motivated by these considerations, we extend the LID task to the subword level (from (a) to (b) in Figure \ref{tab:example1}). 
We introduce a new CS dataset for Spanish--Wixarika (ES--WIX) 
and modify
an existing German--Turkish (DE--TR) CS corpus \cite{cetinoglu:2016a}
for our purposes.
We then introduce a segmental recurrent neural network (SegRNN) model for the task, which we compare against several strong baselines. Our experiments show clear advantages of SegRNNs over all baselines for intra-word CS.



\section{Related Work}
\label{sec:related}
The task of LID for CS has been frequently studied in the last years 
\cite{al2016lili,rijhwani2017estimating,zhang:2018}, 
including two shared tasks on the topic \cite{solorio2014overview,molina2016overview}.
The best systems \cite{samih2016multilingual, rouzbeh2016} 
achieved over $90\%$ accuracy for all language pairs. However, intra-word CS was not handled explicitly, and often systems even failed to correctly assign the \texttt{mixed} label. For Nepali--English, \newcite{barman2014code} correctly identified some of the mixed words with a combination of linear kernel support vector machines and a $k$-nearest neighbour approach. 
The most similar work to ours is \newcite{nguyen2016}, which focused on detecting intra-word CS for Dutch--Limburgish \cite{nguyen2015audience}. The authors utilized Morfessor \cite{creutz2002} to segment all words into morphemes and Wikipedia to assign LID probabilities to each morpheme. However, their task definition and evaluation are on the word level. 
{\color{ForestGreen}Furthermore, as this method relies on large monolingual resources, it is not applicable to low-resource languages like Wixarika, which does not even have its own Wikipedia edition.}


Subword-level LID consists of both segmentation and tagging of words. An earlier approach to handle a similar scenario was the connectionist temporal classification (CTC) model developed by \newcite{graves2006connectionist}. 
The disadvantage of this model was the lack of prediction of the segmentation boundaries that are necessary for our task. \newcite{kong2016segmental} later proposed the SegRNN model that segments and labels jointly, with successful applications on automatic glossing of polysynthetic languages \cite{micher2017improving, micher2018using}. Segmentation of words into morphemes alone has a long history in NLP \cite{harris1951methods}, including semi- or unsupervised methods \cite{goldsmith2001unsupervised, creutz2002, hammarstrom2011unsupervised, gronroos2014morfessor}, as well as supervised ones \cite{zhang2008joint, ruokolainen2013supervised, cotterell2015labeled, kann2018fortification}.

\section{Task and Data Description}
\label{sec:taskandata}


\subsection{Task Description}
\label{subsec:task}

Formally, the task of subword-level LID consists of producing two sequences, given an input sequence of tokens $X=\langle x_1, \dots, x_i, \dots, x_{|X|} \rangle$. The first sequence contains all words and splits $X^s=\langle x^s_1, \dots, x^s_i, \dots, x^s_{|X|}  \rangle$, where each $x^s_i$ is an $m$-tuple of variable length $0 < m \leq |x_i|$, where $|x_i|$ is the number of characters in $x_i$. The second sequence is such that $T^s=\langle t^s_1, \dots, t^s_i, \dots, t^s_{|X|} \rangle$, where $|T^s| = |X^s| = |X|$ and each $t^s_i \in T^s$ is an $n$-tuple of tags from a given set of LID tags. An input--output example for a DE--TR mixed phrase is shown in Figure \ref{fig:example2}.

\begin{figure}[!htbp]
    \centering
    \setlength{\tabcolsep}{1.2pt}
    \small
\begin{tabular}{c|l l l l l}
     Input & $\langle$ `Yerim', &`seni',& `,',& `danke',& `Schatzym'$\rangle$\\\hline
     Output& $\langle$ (Yerim), & (seni), & (,), & (danke), & (Schatzy, m)$\rangle$\\
           & $\langle$ (TR), & (TR), &(OTHER), &(DE), &(DE, TR)$\rangle$
\end{tabular}
    \caption{Subword-level LID in German--Turkish.}
    \label{fig:example2}
\end{figure}

\subsection{Datasets}
\label{subsec:data}
\paragraph{German--Turkish}
The German--Turkish Twitter Corpus \citep{cetinoglu:2016b} consists of 1029 tweets with 17K tokens. They are manually normalized, tokenized, and annotated with language IDs. The language ID tag set consists of \texttt{TR} (Turkish), \texttt{DE} (German), \texttt{LANG3} (other language),  \texttt{MIXED} (intra-word CS), \texttt{AMBIG} (ambiguous language ID in context), and \texttt{OTHER} (punctuation, numbers, emoticns, symbols, etc.). Named entities are tagged with a combination of \texttt{NE} and their language ID: \texttt{NE.TR}, \texttt{NE.DE}, \texttt{NE.LANG3}. In the original corpus, some Turkish and mixed words undergo a morphosyntactic split,\footnote{E.g., separating copular suffixes from roots they are attached to, cf. \citet{cetinoglu:2016b} for details.} with splitting points not usually corresponding to language boundaries. For the purpose of subword-level LID, these morphosyntactic splits are merged back into single words. We then manually segment \texttt{MIXED} words at language boundaries, and replace their labels with more fine-grained language ID tags.
The total percentage of mixed words is $2.75\%$.
However, the percentage of sentences with mixed words is 15.66\%. {\color{ForestGreen}The complete dataset statistics 
can be found in Table \ref{tab:data_trde}}. 
\ozlem{typo}\manuel{corrected}
\ozlem{Correct the Table 1 caption, adapt it to the Table 2 caption. Corrections: LIDs -> language IDs, The All -> All}
\manuel{Done.}


\begin{table}[!htbp]
    \centering
     \small
    \begin{tabular}{l ||r r |r r}
        Tokens & All & \% & Unique & Unique \% \\ \hline
        DE & 3992 & 20.37 & 1360 & 20.43 \\
        TR & 9913 & 50.59 & 4071 & 61.16 \\
        LANG3 & 112 & 0.57 & 83 & 1.25 \\
        AMBIG & 32 & 0.16 & 23 & 0.18 \\
        OTHER & 4345 & 22.17 & 294 & 4.42 \\
        NE.TR & 417 & 2.13 & 275 & 4.13 \\
        NE.DE & 389 & 1.99 & 244 & 3.67 \\
        NE.AMBIG & 16 & 0.08 & 12 & 1.25\\
        NE.LANG3 & 112 & 0.57& 95 & 1.43 \\
        MIXED & 231 & 1.18 & 183 & 2.75 \\
        \hline
        \quad \textit{DE TR} & 231 & 100.0 & 183 & 100.0 \\
        
    \end{tabular}
    \caption{The frequency breakdown of tokens by language IDs in the German-Turkish dataset. \textit{All}: the total number of tokens per tag, \textit{\%}: the percentage of them with respect to the total number of tokens; \textit{Unique}: the number of unique word types, and \textit{Unique \%}: the percentage of them with respect the total number of unique word types.
    }
    \label{tab:data_trde}
\end{table}

\paragraph{Spanish--Wixarika}
Our second dataset consists of 985 sentences and 8K tokens 
in Spanish and Wixarika.
Wixarika is spoken by approximately $50,000$ people in the Mexican states of Durango, Jalisco, Nayarit and Zacatecas \cite{leza2006gramatica} 
and is polysynthetic, with most morphemes ocurring in verbs.
The data is collected from public postings and comments from Facebook accounts. To ensure the public characteristic of these posts, we manually collect data that is accessible publicly 
without being logged in to Facebook, to comply with the terms of use and privacy of the users. These posts and comments are taken from 34 users: 14 women, 10 men, and the rest does not publically reveal their gender. None of them have publically mentioned their age. To get a dataset that focuses on the LID task, we only consider threads where the CS phenomenon appears. 
We replace usernames with \texttt{@username} in order to preserve privacy. Afterwards, we tokenize the text, segment mixed words, and add language IDs to words and segments. 
 
 \begin{table}[!htbp]
    \centering
     \small
    \begin{tabular}{l ||r r |r r}
        Tokens & All & \% & Unique & Unique \% \\ \hline
        ES & 4218 & 50.73 & 1527 &  45.76\\
        WIX & 2019 & 24.28 & 1191 & 35.69 \\
        EN    & 24 & 0.29 & 21 & 0.63 \\
        AMBIG & 28 & 0.34 & 25 & 0.75 \\
        OTHER& 1664 & 20.01 & 288 & 8.63 \\
        NE.ES & 96 & 01.15 & 85 & 2.55 \\
        NE.WIX & 77 & 0.93 & 49 & 1.47 \\
        NE.EN&  11 & 0.13 &9  & 0.27 \\
        MIXED & 177 & 02.13 & 142 & 4.26 \\ \hline
        \quad \textit{ES WIX} & 35 & \textit{19.77}& 31 & \textit{21.83} \\ 
        \quad \textit{WIX ES} & 122 & \textit{68.93}& 93 & \textit{65.49} \\ 
        \quad \textit{WIX ES WIX} & 17 & \textit{9.60}& 31 & \textit{10.56} \\ 
        \quad \textit{WIX EN} & 1 & \textit{0.07} & 1 & \textit{0.07} \\ 
        \quad \textit{EN ES} & 1 & \textit{0.07} & 1 & \textit{0.07} \\ 
        
    \end{tabular}
    \caption{Number of tokens classified by language tags seen in the Spanish-Wixarika dataset. We show the total number of \textit{Tokens} per tag, their proportion (\textit{\%}) with the total tokens, the \textit{Unique} word types, and the proportion (\textit{U. \%}) of them with the total number of unique word types.  }
    \label{tab:data_wixes}
\end{table}
 
The tag set is parallel to that of German--Turkish: \texttt{ES} (Spanish), \texttt{WIX} (Wixarika), \texttt{EN} (English), \texttt{AMBIG} (ambiguous) 
\texttt{OTHER} (punctuation, numbers, emoticons, etc), \texttt{NE.ES}, \texttt{NE.WIX} and \texttt{NE.EN} (named entities). Mixed words are segmented and each segment is labeled with its corresponding language (\texttt{ES}, \texttt{WIX}, \texttt{EN}).
{\color{ForestGreen}Table \ref{tab:data_wixes} shows a detailed description of the dataset.} The percentage of mixed words is higher than in the DE--TR dataset:  $3.13\%$ of the tokens and $4.26\%$ of the types. The most common combination is Spanish roots with Wixarika affixes. Furthermore, 16.55\% of the sentences contain mixed words.
\ozlem{the URL is empty at the moment!}
\manuel{Thank you! :)}


We split the DE--TR corpus and the ES--WIX corpus into training and test sets of sizes 800:229 and 770:216, respectively. Error analysis and hyperparameter tuning are done on the training set via 5-fold cross-validation. We present results on the test sets. Both datasets are  available at \url{https://www.ims.uni-stuttgart.de/institut/mitarbeiter/ozlem/NAACL2019.html}


\begin{table*}[!ht]
    \setlength{\tabcolsep}{5pt} 
  \small
  \centering
  \begin{tabular}{l || r  r  r | r  r  r | r || r  r  r | r  r  r | r} 
    & \multicolumn{7}{c ||}{DE--TR} & \multicolumn{7}{c}{ES--WIX} \\ \hline
    & \multicolumn{3}{c |}{Segmentation} & \multicolumn{3}{c |}{Tagging} & Char & \multicolumn{3}{c |}{Segmentation} & \multicolumn{3}{c |}{Tagging} & Char\\ 
    & P & R & F1 & P & R & F1 &Acc. & P & R & F1 &  P & R & F1 & Acc. \\\hline 
    SegRNN    & 60.4 & 46.8 & \textbf{53.0} & 78.8 & 60.2 & 74.0 & \textbf{72.9}         & 75.6 & 62.7 & \textbf{68.5} & 85.3 & 70.5 & \textbf{77.2} & \textbf{84.6}\\ %
    BiLSTM+\StoS   & 46.1 & 33.4 & 38.7 & 84.3 & 66.8 & \textbf{74.5} & 67.7         & 66.6 & 52.2 & 58.5 & 82.4 & 66.2 & 73.4 &78.4  \\%
    BiLSTM+CRF   & 49.4 & 34.5 & 40.6 & 84.3 & 66.8 & \textbf{74.5} & 68.0         & 61.1 & 48.4 & 54.0 & 82.4 & 66.3 & 73.4 &76.7 \\%
    CRFTag+\StoS     & 12.7 & 6.8 & 8.9 & 27.8 & 15.2 & 19.7          & 37.2         & 47.2 & 31.9 & 38.1 & 63.5 & 43.0 & 51.3 &69.6 \\%
    CRFTag+CRF     & 11.0 & 5.9 & 7.7 & 27.8 & 15.2 & 19.7          & 36.8         & 47.6 & 32.4 & 38.6 & 63.5 & 43.0 & 51.3 &69.4  \\%
    CharBiLSTM & 19.1 & 26.6 & 22.2 & 32.9 & 45.7 & 38.2          & 61.1         & 49.7 & 52.2 & 50.8 & 63.1 & 68.1 & 66.5 &75.7 \\%
  \end{tabular}
  \caption{Segmentation and LID test results for {mixed words only}.\label{results_2}}
\end{table*} 
\ozlem{we sometimes say LID, sometimes tagging. should make them consistent in the text and in the tables}
\manuel{Should I replace it with `LID tagging'? Or only `LID'?}
\ozlem{In our paper LID stands for Language Identification. So Language Identification Tagging doesn't make sense. If 'LID' is used as Language ID we should explicitly write it, e.g. Table 1 caption (there could be some others, I didn't check)}
\manuel{Oh! I see. I will read it and try to find such cases.}

\section{Experiments}
\label{sec:experiments}
Our main system is a neural architecture that jointly solves the segmentation and language identification tasks. We compare it to multiple pipeline systems and another joint system. 
\subsection{SegRNN}
\label{sec:segrnn}
We suggest a SegRNN \cite{kong2016segmental} would be the best fit for our task because
it models a joint probability distribution 
over possible segmentations of the input and labels for each segment.

The model is trained to optimize the following objective, which corresponds to the joint log-likelihood of the segment {\color{ForestGreen}lengths} $e$ and the language tags $t$:
\begin{align}
  {\cal L(\theta)}\!=& \!\!\!\sum_{(x, t, e) \in {\cal D}}  -  \log  p(t,e |x) \label{eq:ll} 
\end{align}
 ${\cal D}$ denotes the training data, $\theta$ is the set of model parameters, $x$ the input, $t$ the tag sequence and $e$ the sequence of segment {\color{ForestGreen}lengths}. 
 
Our inputs are single words.\footnote{We also experimented with entire phrases as inputs, and the achieved scores were slightly worse than for word-based inputs.} As hyperparameters we use: 1 RNN layer, a 64-dimensional input layer, 32 dimensions for tags, 16 for segments, and 4 for {\color{ForestGreen}lengths}. For training, we use Adam \cite{kingma2014adam}. 

%

\subsection{Baselines}
\label{subsec:baselines}


\paragraph{BiLSTM+\StoS/BiLSTM+CRF} Our first baselines are pipelines. First, the input text is tagged with language IDs. 
Language IDs of a mixed word are directly predicted as a combination of all language ID tags of the word (i.e., \texttt{WIX\_ES}). Second, a subword-level model segments words with composed language ID tags. 
For word-level tagging, we use a hierarchical bidirectional LSTM (\textbf{BiLSTM}) that incorporates both token- and character-level information \cite{plank2016multilingual}, similar to the winning system \cite{samih2016multilingual} of the Second Code-Switching Shared Task \cite{molina2016overview}. \footnote{{\color{ForestGreen}For all BiLSTM models input dimension is 100 with a hidden layer size of 100. For training we use a stochastic gradient descent \cite{bottou2010large}, 30 epochs, with a learning rate of 0.1. A 0.25 dropout factor is applied.}} \manuel{Added hyperparameters to this baseline as requested by the reviewers. Not sure if we should keep it.}
For the subword level, we use two 
supervised segmentation methods:
a \textbf{CRF} segmenter proposed by \newcite{ruokolainen2013supervised}, that models segmentation as a labeling problem and a sequence-to-sequence (\textbf{\StoS}) model trained with an auxiliary task as proposed by \newcite{kann2018fortification}.

\paragraph{CRFTag+\StoS/CRFTag+CRF} Since our datasets might be small for training neural networks, we substitute the BiLSTM with a CRF tagger \cite[\textbf{CRFTag}]{muller2013efficient} in the first step. For segmentation, we use the same two approaches as for the previous baselines.

\paragraph{CharBiLSTM} 
We further employ a BiLSTM to tag each character with a language ID. For training, each character inherits the language ID of the word or segment it belongs to. At prediction time, if the characters of a word have different language IDs, the word is split. 

\subsection{Metrics}
\label{subsec:metrics}
We use two 
metrics for evaluation. First, we follow \newcite{kong2016segmental} and calculate precision (P), recall (R), and F1, using segments as units (an unsegmented word corresponds to one segment). 
We also report a tagging accuracy (Char Acc.) by assigning a language ID to each character and calculating the ratio of correct language tags over all characters.

\begin{table}[!htbp]
    \setlength{\tabcolsep}{3.pt} 
  \small
  \centering
  \begin{tabular}{l | c  c  c  | c  c  c} 
    & \multicolumn{3}{c |}{DE--TR} & \multicolumn{3}{c}{ES--WIX} \\ \hline
    & Seg. & Tag. & Char  & Seg.& Tag. & Char\\ 
    &  F1 &  F1 & Acc.&  F1 & F1  & Acc. \\\hline %
    SegRNN       & \textbf{98.7} &         94.0  &        93.6 &         97.8 & \textbf{92.5}&  \textbf{92.4}\\ 
    BiLSTM+\StoS     &         98.6  & \textbf{95.1} &         94.3 & \textbf{98.1}&         90.9 &          90.7  \\%
    BiLSTM+CRF     &         \textbf{98.7}  &         94.9  & \textbf{94.4}&         97.9 &         87.8 &          90.6   \\%
    CRFTag+\StoS        &         98.4  &         93.7  &         93.1 &         97.7 &         90.4 &          90.1 \\%
    CRFTag+CRF        &         98.4  &         93.7  &         93.1 &         97.6 &         90.3 &          90.1 \\%
    CharBiLSTM   & 87.7 &  88.0 &  92.5&         89.7  &         87.9  &         91.3  \\%
  \end{tabular}
  \caption{Test set results for {entire datasets}.\label{results_1}}
\end{table}

\subsection{Results and Discussion}
\label{subsec:results}
Table \ref{results_1} 
shows all test results for the entire datasets. We find the following: (i) For ES--WIX,
SegRNN performs slightly better for tagging than the best baseline, both in terms of F1 and character accuracy. 
For DE--TR, SegRNN and BiLSTM+CRF are the best segmentation models, but the BiLSTM models slightly outperform SegRNN for tagging.
(ii) The CRF pipelines perform slightly worse than the best word-level BiLSTM models for both datasets and all evaluations.

\begin{figure*}[htbp]
    \centering
    \includegraphics[width=1\textwidth]{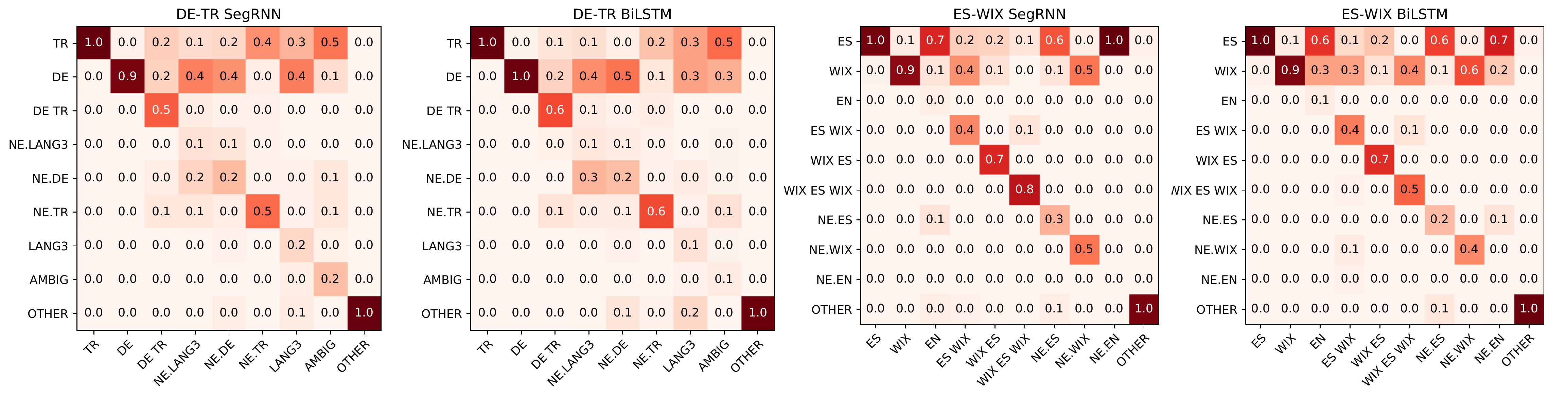}
    \caption{{\color{ForestGreen}Confusion matrices of the two best models on both datasets. The $x$ axis represents tags seen in the gold standard, and the $y$ axis shows the corresponding predicted tags. Values are rounded up, therefore not all columns add up to 1.}}
    \label{fig:confiusion_matrix}
\end{figure*}

Table \ref{results_2} shows the results of tagging and segmentation only for the mixed words in our datasets. Here, we can see that: (i) Our SegRNN model achieves the best performance for segmentation. Differences to the other approaches are  $\geq10\%$, showing clearly why these models are good for the task when the number of words belonging to two languages is high. (ii) The pipeline BiLSTM models work best for tagging of the DE--TR data with a slight margin, but underperform on the ES--WIX dataset as compared to the SegRNN models. (iii) Both CRFTag models achieve very low results for both segmentation and tagging. (iv) CharBiLSTM performs better than the CRFTag models on both tasks, but is worse than all other approaches in our experiments.

More generally, we further observe that recall on mixed words for the DE--TR pair is low for all systems, as compared to ES--WIX. This effect is especially strong for the CRFTag and CharBiLSTM models, which seem to be unable to correctly identify mixed words. While this tendency can also be seen for the ES--WIX pair, it is less extreme. 
We suggest that the better segmentation and tagging of mixed words for ES--WIX might mostly be due to 
the higher percentage of available examples of mixed words in the training set for ES--WIX.

Overall, we conclude that SegRNN models seem to work better on language pairs that have more intra-word CS, while pipeline approaches might be as good for language pairs where the number of mixed words is lower.

\ozlem{fig3 caption: Values are rounded up, therefore not all columns add up to 1. I don't repeat my other comments about the matrices.}

\paragraph{Error analysis.}{\color{ForestGreen} Figure \ref{fig:confiusion_matrix} \katharina{The caption is grammatically incorrect.}\manuel{changed sums to sum. Is there other grammar error?} shows confusion matrices for SegRNN and BiLSTM+Seg2Seg. Both models achieve good results assigning monolingual tags (\texttt{ES}, \texttt{WIX}, \texttt{DE}, \texttt{TR}) and punctuation symbols (\texttt{OTHERS}). 
The hardest labels to classify are named entities (\texttt{NE}, \texttt{NE.TR}, \texttt{NE.TR}, \texttt{NE.WIX}, \texttt{NE.ES}), as well as third language and ambiguous tags (\texttt{LANG3}, \texttt{EN}, \texttt{AMBIG}). Performance on multilingual tags (\texttt{DE TR}, \texttt{WIX ES}, \texttt{ES WIX}, \texttt{WIX ES WIX}) is mixed. For \texttt{DE TR}, BiLSTM+Seq2Seq gets slightly better classifications, but for the ES--WIX tags SegRNN achieves better results. \katharina{Write tags in the same way always!} \manuel{ups!}
\katharina{Still wrong.}

Regarding \ozlem{in regards to -> regarding}\manuel{Done} oversegmentation problems, BiLSTM+Seq2Seq (0.8\% for DE--TR and 2.0\% for ES--WIX) slightly underperforms \ozlem{under performs -> underperforms}\manuel{done} SegRNN (0.7\% for DE--TR and 1.13\% for ES--WIX). The BiLSTM+Seq2Seq (2.4\%) makes fewer undersegmentation errors for DE--TR than SegRNN (2.7\%). However, for ES--WIX, SegRNN performs better with 3.81\% undersegmentation errors compared to 4.2\% of BiLSTM+Seq2Seq. 

}

\section{Conclusion}
\label{sec:conclusion}
{\color{ForestGreen} In this paper,} we extended the LID task to the subword level, which is
particularly important for code-switched text in morphologically rich languages. 
We further proposed a SegRNN model for the task and compared it to several strong baselines. Investigating the behaviour of all systems, 
we found that pipelines including a BiLSTM tagger
work well for tagging DE--TR, where the number of mixed tokens is not that high, but that our proposed SegRNN approach performs better than all other systems for ES--WIX.
Also, SegRNNs have clear advantages over all baselines if we consider mixed words only. Our subword-level LID datasets for ES--WIX and DE--TR are publicly available.
\katharina{IMPORTANT: change that sentence!!}
\manuel{Deleted. We already have a footnote with a link to the datasets.}
\katharina{Please repeat the mention of the dataset in the conclusion.}

\section*{Acknowledgments}
{\color{ForestGreen}
We would like to thank Mohamed Balabel, Sam Bowman,  Agnieszka Falenska, Ilya Kulikov and Phu Mon Htut for their valuable feedback. We also want to thank Jeffrey Micher for his help with the setup of the SegRNN code. 
This project has benefited from financial support to Manuel Mager and \"{O}zlem \c{C}etino\u{g}lu by DFG via project CE 326/1-1 ``Computational \underline{S}tructural \underline{A}nalysis of \underline{G}erman-\underline{T}urkish Code-Switching'', to Manuel Mager by DAAD Doctoral Research Grant, and to Katharina Kann by Samsung Research.

}
\ozlem{:) Initials are unnecessary when there are still 5-6 empty lines. When used, my initials should have diacritics.}
\bibliography{naaclhlt2019}
\bibliographystyle{acl_natbib}

\end{document}